\documentclass{article}

\usepackage{amsmath}
\usepackage{booktabs}
\usepackage{amssymb}
\usepackage[dvipsnames]{xcolor}
\usepackage{tabularx}
\usepackage{multirow} 
\usepackage{makecell} 
\usepackage{colortbl} 
\usepackage[Export]{adjustbox} 
\usepackage{arydshln} 
\usepackage{colortbl}
\usepackage{pifont} 
\usepackage{tikz} 
\usepackage{bm}
\usepackage{caption}
\usepackage{subcaption} 
\usepackage{array}

\colorlet{colorFst}{Green!25}       
\colorlet{colorSnd}{SpringGreen!45} 
\colorlet{colorTrd}{Yellow!30}      
\colorlet{colorLow}{darkgray!30}    

\definecolor{tabfirst}{rgb}{1, 0.7, 0.7} 
\definecolor{tabsecond}{rgb}{1, 0.85, 0.7} 
\definecolor{tabthird}{rgb}{1, 1, 0.7} 

\newcommand{\fs}{\cellcolor{tabfirst}\bf}   
\newcommand{\nd}{\cellcolor{tabsecond}}      
\newcommand{\rd}{\cellcolor{tabthird}}      

\newcolumntype{R}{>{\raggedleft\arraybackslash}X}

\usepackage{tocloft}

\usepackage{PRIMEarxiv}

\usepackage[utf8]{inputenc} 
\usepackage[T1]{fontenc}    
\usepackage{hyperref}       
\usepackage{url}            
\usepackage{booktabs}       
\usepackage{amsfonts}       
\usepackage{nicefrac}       
\usepackage{microtype}      
\usepackage{lipsum}
\usepackage{fancyhdr}       
\usepackage{graphicx}       
\graphicspath{{media/}}     

\title{GSFF-SLAM: 3D Semantic Gaussian Splatting SLAM via Feature Field
}
\author{
  Zuxing Lu, Xin Yuan, Shaowen Yang,  Jingyu Liu, Changyin Sun\thanks{Corresponding author}  \\
  Southeast University \\
  Nanjing \\
  \texttt{\{luzuxing, xinyuan, 220232017, 230239497, cysun\}@seu.edu.cn} \\
   \And
  Jiawei Wang, \\
  Tongji University\\
  Shanghai \\
  \texttt{wangjw@tongji.edu.cn} \\
}

\begin{document}
\maketitle

\begin{abstract}



Semantic-aware 3D scene reconstruction is essential for autonomous robots to perform complex interactions. Semantic SLAM, an online approach, integrates pose tracking, geometric reconstruction, and semantic mapping into a unified framework, shows significant potential. However, existing systems, which rely on 2D ground truth priors for supervision, are often limited by the sparsity and noise of these signals in real-world environments.
To address this challenge, we propose GSFF-SLAM, a novel dense semantic SLAM system based on 3D Gaussian Splatting that leverages feature fields to achieve joint rendering of appearance, geometry, and N-dimensional semantic features. By independently optimizing feature gradients, our method supports semantic reconstruction using various forms of 2D priors, particularly sparse and noisy signals. Experimental results demonstrate that our approach outperforms previous methods in both tracking accuracy and photorealistic rendering quality. When utilizing 2D ground truth priors, GSFF-SLAM achieves state-of-the-art semantic segmentation performance with 95.03\% mIoU, while achieving up to 2.9$\times$ speedup with only marginal performance degradation.

\end{abstract}
\section{Introduction}

Visual Simultaneous Localization and Mapping (Visual SLAM) is a widely used technique in robotics and computer vision~\cite{davison2007monoslam,fuentes2015visual,klein2007parallel,mur2015orb,mur2017orb,newcombe2011dtam}, enabling intelligent agents to reconstruct 3D scenes of unknown environments using monocular or multiple cameras while simultaneously tracking their positions over time.
Neural Radiance Fields (NeRF)~\cite{mildenhall2021nerf} and 3D Gaussian Splatting (3DGS)~\cite{kerbl20233d} are two recently emerging 3D reconstruction methods. Notably, their high-quality 3D geometric representations and novel view synthesis capabilities significantly enhance the utilization efficiency of high-dimensional image inputs. These advancements have facilitated the transition from traditional sparse Visual SLAM to learnable dense Visual SLAM systems~\cite{zhu2022nice,yang2022vox,sandstrom2023point,keetha2024splatam,matsuki2024gaussian}.

\begin{figure}[t]
    \centering
    \includegraphics[width=0.7\linewidth]{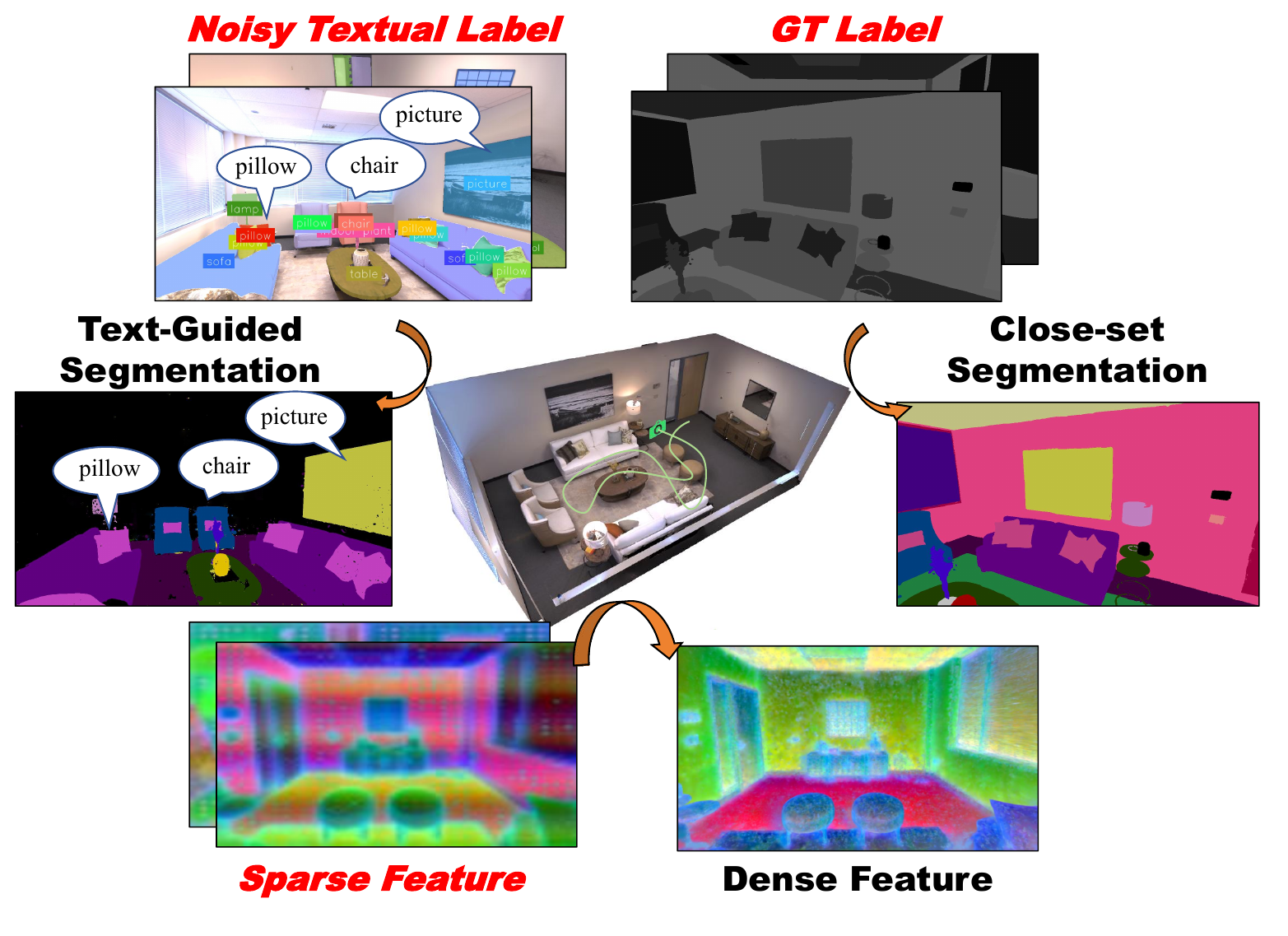}
    \caption{Our GSFF-SLAM leverages different forms of signals to enhance various downstream online tasks. Our method projects 2D priors into the 3D feature field, enabling high-precision close-set segmentation, text-guided segmentation, and dense feature map rendering.}
    \label{fig:downstream tasks}
\end{figure}

Semantic SLAM aims to maintain robust tracking accuracy over long sequences without relying on precise Structure-from-Motion (SFM)~\cite{schoenberger2016sfm} poses, and it performs reconstruction from sparse multi-view frames while compressing dense semantic information into 3D scene representations. For instance, SemanticFusion~\cite{mccormac2017semanticfusion} associates each point in the point cloud with a semantic label, thereby enriching the scene representations. 
Inspired by offline reconstruction works~\cite{zhi2021place,kobayashi2022decomposing}, NeRF-based semantic SLAM methods~\cite{haghighi2023neural,li2024dns,zhu2024sni} represent scenes using the learnable implicit neural networks and achieve semantic map rendering by integrating additional semantic Multi-Layer Perceptrons (MLPs). However, these methods are more prone to catastrophic forgetting than offline approaches, especially in long sequences, due to the absence of memory replay mechanisms.
In contrast, the SGS-SLAM~\cite{li2024sgs} avoids this issue by utilizing explicit 3D Gaussians as the scene representation, and it achieves high-speed semantic map rendering by converting semantic labels into RGB images. 
These Semantic SLAM methods~\cite{haghighi2023neural,li2024dns,zhu2024sni,li2024sgs} integrate semantic losses into tracking and mapping, improving localization accuracy with multi-view consistent priors such as ground truth annotations. However, this design relies heavily on high-quality priors, restricting its use in scenarios lacking such annotations.
Moreover, the effectiveness of semantic SLAM methods utilizing noisy and sparse priors remains unverified.

To address these limitations, we propose a novel approach that represents the semantic information through N-dimensional feature fields. 
We decouple the semantic optimization process from scene reconstruction by independently optimizing semantic embedding gradients. Specifically, we first reconstruct the input images to obtain high-quality geometric representations before performing semantic optimization. This decoupling of the mapping process into two steps enables the framework to support diverse forms of supervision signals.
As demonstrated in the text-guided segmentation results shown in Figure~\ref{fig:downstream tasks}, our method achieves high-quality semantic reconstruction even with noisy and sparse textual priors. Furthermore, the feature field densifies sparse features extracted from foundation models, enriching each 3D Gaussian in the scene with comprehensive semantic information.
Overall, our contributions are summarized as follows:

\begin{itemize}
    \item We propose GSFF-SLAM, a novel Semantic SLAM framework based on 3D Gaussian Splatting, which leverages N-dimensional feature fields to achieve high-quality semantic reconstruction and dense feature map rendering.
    \item Our framework decouples semantic optimization from scene reconstruction by independently optimizing semantic embedding gradients, ensuring robust performance even with noisy and sparse 2D priors.
    \item On the Replica dataset~\cite{straub2019replica}, our method achieves state-of-the-art semantic segmentation performance of 95.03\% mIoU with 2D ground truth priors, while delivering a runtime improvement of up to 2.9$\times$ with only a slight trade-off in performance.
\end{itemize}

\section{Related Work}

\textbf{3D Gaussian Splatting.}
3DGS~\cite{kerbl20233d} is an explicit 3D scene representation that introduces learned 3D Gaussians, $\alpha$-blending, and an efficient parallel Gaussian rasterizer.
Benefiting from its high rendering speed and explicit representation capabilities, 3DGS has facilitated significant advancements in various applications, including Visual SLAM~\cite{keetha2024splatam, matsuki2024gaussian, yugay2023gaussian}, language-guided scene editing~\cite{chen2024gaussianeditor} and offline semantic reconstruction~\cite{shi2024language, zhou2024feature}.
MonoGS~\cite{matsuki2024gaussian} proposes a monocular SLAM system based on 3DGS, which selects keyframes based on inter-frame co-visibility instead of fixed frame intervals or predefined distance-angle thresholds.
LEGaussians~\cite{shi2024language} introduces semantic feature embedding into 3DGS, enabling offline semantic reconstruction.
GaussianEditor~\cite{chen2024gaussianeditor} utilizes the SAM~\cite{kirillov2023segany} model to segment and annotate semantics in 2D images, followed by explicit semantic information retrieval and editing.

\noindent \textbf{Dense Visual SLAM.}
SLAM typically divided into two main tasks: mapping and tracking~\cite{klein2009parallel}.
Unlike sparse SLAM methods that primarily focus on pose estimation~\cite{campos2021orb, mur2015orb, mur2017orb}, dense visual SLAM methods aim to reconstruct detailed 3D maps~\cite{sucar2021imap, yang2022vox,zhu2022nice}.
Map representations in dense SLAM can be broadly categorized into two types: frame-centric and map-centric. 
Frame-centric methods anchor 3D geometry to specific keyframes, estimating frame depth and inter-frame motion~\cite{czarnowski2020deepfactors,newcombe2011dtam}. On the other hand, Map-centric methods convert 2D images into 3D geometry aligned with a unified world coordinate system.
Common 3D geometry primitives include point clouds~\cite{cao2018real,sandstrom2023point,teed2021droid,zhou2013elastic}, voxel grids~\cite{dai2017bundlefusion, muglikar2020voxel,yang2022vox,zhu2022nice}, and 3D Gaussians~\cite{keetha2024splatam,matsuki2024gaussian,yugay2023gaussian}. 
Point clouds can flexibly adjust the sparsity of spatial points, but due to the lack of correlation between primitives, the design of optimization is more challenging. 
Voxel grids enable fast 3D retrieval but incur high memory and computational costs. The NeRF-based SLAM methods~\cite{sucar2021imap, zhu2022nice} significantly reduce spatial occupancy by converting explicit voxel grids into neural implicit representations. As a novel primitive, 3D Gaussians exhibit differentiable and continuous properties, enabling 3DGS-based SLAM methods~\cite{keetha2024splatam, matsuki2024gaussian, yugay2023gaussian} to achieve efficient training and inference.

\noindent\textbf{Semantic Reconstruction.}
Semantic reconstruction focuses on enriching occupancy maps by integrating semantic information into map structures~\cite{haghighi2023neural,huang2023visual,kobayashi2022decomposing,shen2023distilled,wei2024ovexp,zhi2021place,zhou2024feature,zhu2024sni}.
2D semantic reconstruction methods typically fuse 2D priors with the map through projection, generating top-down multi-layer map structures~\cite{huang2023visual,wei2024ovexp}. These methods emphasize interaction with complex natural language, generating one or more trajectories, with a focus on language-guided planning capabilities. 
In contrast, 3D methods prioritize precise localization, accurate contours, and edge details~\cite{haghighi2023neural,zhu2024sni}. NeRF-based methods output semantic features using additional MLPs, often sharing parameters with spatial occupancy networks. For instance, Semantic-NeRF~\cite{zhi2021place} embeds noisy and sparse semantic signals into its framework, enabling robust reconstruction of semantic information even under challenging conditions.  NeRF-DFF~\cite{kobayashi2022decomposing} proposes distilling feature fields to achieve zero-shot semantic segmentation.  However, these methods suffer from catastrophic forgetting and local detail loss, necessitating memory replay and hash position encoding~\cite{muller2022instant}.
3DGS-based methods, such as Feature-3DGS~\cite{zhou2024feature}, leverage explicit representations for stable feature binding, distilling teacher models (LSeg~\cite{li2022language} and SAM~\cite{kirillov2023segany}) to achieve up to 2.7$\times$ faster rendering. 
Nevertheless, applying offline methods~\cite{zhi2021place,kobayashi2022decomposing,zhou2024feature} to online tasks remains challenging due to inaccurate pose estimation and insufficient semantic supervision signals. Our method leverage sparse multi-view information to achieve accurate pose estimation and online semantic reconstruction.

\begin{figure*}[!t]
    \centering
    \includegraphics[width=\linewidth]{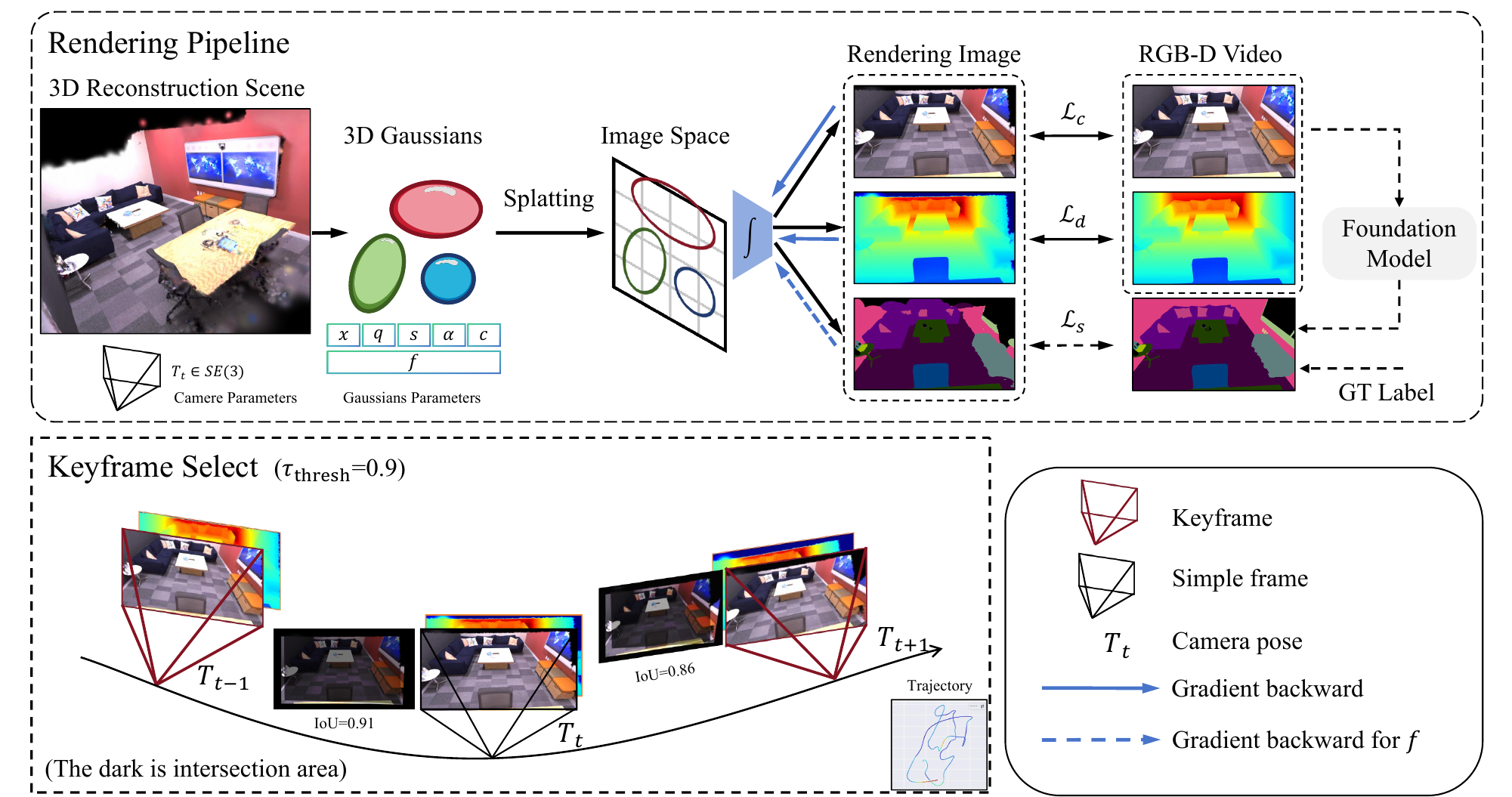}
    \caption{\textbf{Overview of GSFF-SLAM}. Our method takes an RGB-D stream as input, leveraging 3D Gaussian Splatting with semantic feature embedding \(f\) to generate RGB images, depth images, and dense feature maps. Semantic signals, derived from foundation models or ground truth, supervise the learning process, while the feature embedding \(f\) is optimized independently.}
    \label{fig:system framework}
\end{figure*}

\section{Method}

Given an RGB-D video stream in unknown static environments, our goal is to reconstruct the 3D scene structure and semantic information while tracking the camera pose. Figure \ref{fig:system framework} outlines the general framework of GSFF-SLAM.

\subsection{Scene Representation and Rendering}
We utilize isotropic Gaussian points to represent the scene, where the number of 3D Gaussians \(\mathcal{N}\) dynamically adapts during the optimization process. To implement the semantic reconstruction, we introduce a new trainable parameter, the semantic feature embedding \(f \in \mathbb{R}^N\), for each Gaussian point. The optimizable properties of each Gaussian point are given by \(\mathcal{G}_i = \{x_i, q_i, s_i, \alpha_i, c_i, f_i\}\), where \(x \in \mathbb{R}^3\) denotes the mean \(\mu\) (geometric center) and the 3D covariance matrix \(\Sigma\) is expressed:

\begin{equation}
\label{Eq:Sigma_R_S}
\Sigma = RSS^TR^T,
\end{equation}
where \(R\) is the rotation matrix, derived from a quaternion \(q\in\mathbb{R}^4\), and \(S\) is the scaling matrix, constructed from the scaling factor \(s\in\mathbb{R}^3\). Equation \ref{Eq:Sigma_R_S} ensures that the 3D covariance matrix is positive semi-definite and physically meaningful. The opacity value \(\alpha \in \mathbb{R}\) and the third-order spherical harmonics (SH) coefficients \(c \in \mathbb{R}^3\) govern the color of the rendered image. 


We use the rendering pipeline based on the differentiable Gaussian splatting framework proposed in~\cite{kerbl20233d} as the foundation, extending it to render depth images and dense feature maps. These projected 2D Gaussian points are sorted by depth in a front-to-back order and rendered in the camera view using \(\alpha\)-blending:
\begin{equation}
\label{Eq:color_depth_feature}
\begin{aligned}
    \mathbf{c}&=\sum_{i\in\mathcal{N}}c_i\alpha_i \prod_{j=1}^{i-1}(1-\alpha_j), \\
    \mathbf{d}&=\sum_{i\in\mathcal{N}}d_i\alpha_i\prod_{j=1}^{i-1}(1-\alpha_j), 
\end{aligned}
\end{equation}
\begin{equation}
    \mathbf{f} = 
    \begin{cases} 
        \displaystyle \sum_{i \in \mathcal{N}} f_i \alpha_i \prod_{j=1}^{i-1}(1-\alpha_j), & \text{if rendering features}, \\
        0, & \text{otherwise}.
    \end{cases}
\end{equation}

Specifically, during the gradient backward process, our pipeline computes the gradient of the feature \(f\) solely through the supervised semantic maps, without further propagating to the mean \(\mu\) and variance \(\Sigma\). To enhance rendering efficiency, we introduce a parameter that indicates whether to render the feature maps and backpropagate the gradient of the feature embedding \(f\). This design enables the rendering pipeline to achieve high-speed rendering, even during camera pose tracking or scene reconstruction without semantic features.

\subsection{Tracking and Mapping Optimization}

We adopt the tracking strategy of most Visual SLAM methods, which estimates the relative pose changes between consecutive frames and accumulates them to determine the current frame's pose. Additionally, we construct the map by selecting keyframes rather than using all frames, which improves efficiency.  

\noindent\textbf{Camera Pose Optimization.}
To avoid the overhead of automatic differentiation, we follow MonoGS~\cite{matsuki2024gaussian} and compute the gradient of the camera viewpoint transformation matrix directly using Lie algebra, integrating it into the rendering pipeline as shown in Equation \ref{Eq:lie_algebra}:  
\begin{equation}
\label{Eq:lie_algebra}
\frac{d\mu_c}{dT_{CW}} =  [\textit{\textbf{I}}, -\mu_c^{\times}],
\frac{dW}{dT_{CW}} = [ \textit{\textbf{0}},  -W^{\times} ],
\end{equation}
where \({}^{\times}\) denotes the skew symmetric matrix. To ensure stable tracking accuracy, we check the convergence of \(\Delta\mu_c\) and \(\Delta W\) to determine if the current frame has completed tracking.

\noindent\textbf{Keyframe Select.} 
By evaluating the co-visibility, which is the rendering overlap between the current frame and the previous keyframe, we determine whether to add a new keyframe. During rendering, the 3D Gaussian points are sorted based on their distance to the camera, and the transmittance \(T_i=\prod_{j=1}^{i-1}(1-\alpha_j)\) is calculated, defined as the product of the opacity values of the previous Gaussian points in front of the current point along the ray. Gaussian points with a transmittance greater than 0.5 are marked as visible. In detail, we record the visibility \(\mathbf{V_{id}}\) of the rendered Gaussian points during tracking. 
\begin{equation}
    \label{Eq:visibility}
    \mathbf{V_{id}} = \begin{bmatrix}
    v_1 \\
    v_2 \\
    \vdots \\
    v_{\mathcal{N}}
    \end{bmatrix}, v_i \in \{0, 1\}, i = 1, 2, \dots, \mathcal{N},
\end{equation}
where \(\mathbf{id}\) denotes the frame index. We then calculate the intersection-over-union (IoU), and if the IoU is less than the threshold \(\tau_{\text{thresh}}\), the current frame is set as a keyframe. As shown in Figure \ref{fig:system framework}, we record the visibility of all keyframes and use the co-visibility to determine whether to add a new keyframe.

\noindent\textbf{Tracking Loss.} 
During  the tracking process, we first freeze the Gaussian point parameters \(\mathcal{G}\) and then generate the current frame's RGB image and depth image through the rendering pipeline. We also record the 2D visibility map as the mask \(m_{v}\).
Next, we adjust the camera viewpoint \(T_{CW}\) parameters \(\delta\), re-render the images, and iterate until convergence. 
For the RGB image, we compute the image edge gradient to extract edges and retain regions with significant gradients as \(m_{e}\). The tracking loss function is defined as:  
\begin{equation}
    \label{Eq:Tracking}
    \mathcal{L}_{\text{tracking}} = \lambda_{t} m_{v} m_{e}\mathcal{L}_{c}(\delta) + (1-\lambda_{t})m_{v}\mathcal{L}_{d}(\delta),
\end{equation}
where \(\lambda_{t}\) is the hyperparameter of tracking process, and \(\mathcal{L}_{c}\) and \(\mathcal{L}_{d}\) represent the color loss and depth loss with L1 loss, respectively.

\noindent\textbf{Mapping Loss.} For the selected keyframes, we use Open3D~\cite{zhou2018open3d} to initialize the point cloud positions in unseen regions. By downsampling the point cloud density \(\rho_{\text{pc}}\), we control the sparsity of the point cloud to regulate the number of 3D Gaussian points. Then we freeze the camera viewpoint \(T_{CW}\) and optimize the Gaussian point parameters $\mathcal{G}$. By minimizing the difference between the rendered images and the ground truth, we can progressively refine the geometric and color attributes of the Gaussian points. Additionally, to address the issues of Gaussian points being overly elongated in unobserved or growing regions and the uneven sparsity distribution, we introduce a regularization term:
\begin{equation}
    \label{Eq:Reg}
    \mathcal{L}_{\text{reg}} = \frac{1}{\mathcal{N}} \sum_{i=1}^{\mathcal{N}} |s_i - \Bar{s}|_1,
\end{equation}
where \(\Bar{s}\) is the average scaling factor of all Gaussian points. The mapping loss function is defined as:  
\begin{equation}
    \label{Eq:Mapping}
    \mathcal{L}_{\text{mapping}} = \lambda_{m}\mathcal{L}_{c}(\mathcal{G}) + (1-\lambda_{m})\mathcal{L}_{d}(\mathcal{G}) + \lambda_{r}\mathcal{L}_{\text{reg}},
\end{equation}
where \(\lambda_{m}\) is the hyperparameter of mapping process, and \(\lambda_{r}\) regulates the strength of the regularization term.  

\subsection{Feature Field Optimization}

Using the pipeline described in \ref{sec:scene}, we render the dense feature map \(\hat{F} \in \mathbb{R}^{H \times W \times N}\).  Our goal is to bind the feature embedding \(f\) of each Gaussian point to its 3D position through the \textbf{Splatting} process, whereas NeRF-based methods achieve this by sharing part of the network parameters. 

\noindent\textbf{Ground Truth Supervision.}  
For the input supervision signal, the ground truth label \(L \in \mathbb{N}^{H \times W}\), we use the cross-entropy loss function to guide the learning of the dense feature map. The semantic loss is defined as:
\begin{equation}
    \label{gt label}
    \mathcal{L}_s = - \frac{1}{H W} \sum_{i=1}^{H} \sum_{j=1}^{W} \sum_{c=1}^{N} L_{i,j}(c) \log(\hat{F}_{i,j}(c)).
\end{equation}
During evaluation, we use the argmax function to convert \(\hat{F}\) into the predicted label \(\hat{L}\):
\begin{equation}
    \hat{L} = \arg\max(\hat{F}),
\end{equation}

\noindent\textbf{Noisy Textual Label Supervision.}  
To optimize the feature embedding \(f \in \mathbb{R}^{N}\), we minimize the difference between the rendered feature map \(\hat{F} \in \mathbb{R}^{H \times W \times N}\) and the predicted feature map \(I^f \in \mathbb{R}^{H \times W \times M}\). 
Inspired by the concept of distilled feature fields~\cite{kobayashi2022decomposing}, we employ foundation models: Grounding-DINO~\cite{liu2023grounding} for open-vocabulary detection and SAM~\cite{ravi2024sam2} for dense segmentation. Additionally, we use CLIP~\cite{radford2021learning} to encode text queries into feature vectors, with \(M\) typically set to 512.  
Specifically, Grounding-DINO takes an RGB image \(I\) as input and generates a triplet output: bounding boxes \(b\in\mathbb{R}^{K\times4}\), open-vocabulary text labels \(L = \{l_1,l_2,...,l_k\}\), and confidence scores \(s\in\mathbb{R}^{K}\). SAM then processes the bounding boxes \(b\) to produce dense binary segmentation masks \(M\in\{0,1\}^{H \times W}\). The text encoder transforms the text labels into feature vectors, yielding \(I^f \in \mathbb{R}^{H \times W \times M}\). The semantic supervision loss is defined as:
\begin{equation}
\label{semantic_loss}
\mathcal{L}_s=||I^f- o(\hat F)||_1,
\end{equation}
where \(o(\cdot)\) denotes the convolutional upsampling operation.
The probability of each pixel \(x\) in the rendered feature map belonging to a label \(l\) is computed as:  
\begin{equation}
\label{label_sim}
p(l|x)=\frac{\exp{(f(x)q(l)^T)}}{\sum_{l'\in L} \exp{(f(x)q(l')^T)}},
\end{equation}
where \(q(l)\) is the query vector for label \(l\) generated by the text encoder.

The optimization of the feature field is performed after multiple iterations of optimizing the geometric and color attributes. This decoupling is motivated by two key reasons: first, simultaneous optimization significantly increases the computational burden on the optimizer; second, empirical results show that a well-constructed map effectively reduces the iteration count for feature field optimization, enhancing overall efficiency.
\section{Experiments}

\subsection{Datasets, Metrics, and Baselines}

\noindent\textbf{Datasets}. We evaluate our approach on three widely-used datasets. Replica~\cite{straub2019replica} is a high-quality synthetic dataset of 3D indoor environments with detailed geometry and texture. Following previous works, we select 8 scenes and use provided camera poses for experiments. ScanNet~\cite{dai2017scannet}  is a real-world dataset with RGB-D images and semantic annotations. We choose 5 representative scenes to evaluate the detail performance. TUM-RGBD~\cite{sturm2012benchmark} is another real-world dataset with RGB-D images and highly accurate camera poses. We use 5 scenes to evaluate the pose estimation performance.

\noindent\textbf{Metrics}. Our method focuses on two main tasks: SLAM and semantic reconstruction. For SLAM evaluation, We follow metrics from SNI-SLAM~\cite{zhu2024sni}. Specifically, we use \textit{ATE RMSE(cm)} (Absolute Trajectory Error Root Mean Square Error) for tracking accuracy evaluation. For evaluating rendering quality, we employ \textit{PSNR(dB)} (Peak Signal-to-Noise Ratio), \textit{SSIM} (Structural Similarity Index), and \textit{LPIPS} (Learned Perceptual Image Patch Similarity). For semantic reconstruction evaluation, we utilize total pixel accuracy \textit{Acc(\%)} and mean class-wise intersection over union \textit{mIoU(\%)}.

\noindent\textbf{Baselines}. We compare the semantic construction accuracy with state-of-the-art NeRF-based and 3DGS-based methods, including SNI-SLAM~\cite{zhu2024sni}, DNS-SLAM~\cite{li2024dns}, and SGS-SLAM~\cite{li2024sgs}. We also compare the SLAM performance with the other baselines~\cite{zhu2022nice,yang2022vox,sandstrom2023point,keetha2024splatam,matsuki2024gaussian}.

\noindent\textbf{Implementation Details}. We set the feature dimension \(N\) to 128 to represent semantic information. All experiments are conducted on a single NVIDIA RTX 4090 GPU. Please refer to the supplementary material for further details of our implementation.

\subsection{Evaluation of SLAM Metrics}

\begin{table}[t]
    \centering
    \renewcommand{\arraystretch}{1.2}
    \resizebox{0.9\columnwidth}{!}{
    \begin{tabular}{lccccccccc}
    \toprule
    Method                                      &     O0    &     O1    &     O2    &     O3    &     O4    &     R0    &     R1    &     R2    &\textbf{Avg.}\\
    \midrule
    \multicolumn{10}{l}{\cellcolor[HTML]{EEEEEE}{\textit{Neural Implicit Fields}}} \\ 
    \multirow{1}{*}{NICE-SLAM~\cite{zhu2022nice}}         &     0.88 &     1.00 &     1.06 &     1.10 &     1.13 &     0.97 &     1.31 &     1.07 &     1.061  \\[0.8pt]  
    \multirow{1}{*}{Vox-Fusion~\cite{yang2022vox}}        &     8.48 &     2.04 &     2.58 &     1.11 &     2.94 &     1.37 &     4.70 &     1.47 &     3.086 \\[0.8pt]  
    \multirow{1}{*}{DNS-SLAM~\cite{li2024dns}}            &     0.84 &     0.84 &     0.69 &     1.11 &     0.66 &     0.72 &     1.08 &     0.76 &     0.838  \\[0.8pt]  
    \multirow{1}{*}{SNI-SLAM~\cite{zhu2024sni}}           &     0.41 &     0.38 &     0.48 &     0.71 &     0.52 &     0.40 &     0.38 &     0.35 &     0.456  \\[0.8pt]  
    \multirow{1}{*}{Point-SLAM~\cite{sandstrom2023point}} &     0.38 &     0.48 &     0.54 &     0.69 &     0.72 &     0.61 &     0.41 &     0.37 &     0.525  \\[0.8pt]  
    \hdashline 
    \multicolumn{10}{l}{\cellcolor[HTML]{EEEEEE}{\textit{3D Gaussian Splatting}}} \\ 
    \multirow{1}{*}{SplaTAM~\cite{keetha2024splatam}}     &     0.47 &     0.27 &     0.29 &     0.32 &     0.72 &     0.31 &    0.40  &     0.29 &     0.384  \\[0.8pt]  
    \multirow{1}{*}{MonoGS~\cite{matsuki2024gaussian}}    &     0.36 &     0.19 &     0.25 &     0.12 &     0.81 &     0.33 &     0.22 &     0.29 &     0.321  \\[0.8pt]  
    \multirow{1}{*}{SGS-SLAM~\cite{li2024sgs}}            &     0.44 &     0.41 &     0.52 &     0.46 &     0.43 &     0.32 &    0.39  &     0.37 &     0.418  \\[0.8pt]  
    \multirow{1}{*}{GSFF-SLAM(Ours)}                      &     0.35 &     0.24 &     0.25 &     0.15 &     0.54 &     0.34 &     0.29 &     0.33 &     0.311  \\[0.8pt]  
    \bottomrule
    \end{tabular}
    }
    \caption{Comparison of tracking performance on Replica dataset~\cite{straub2019replica}. We report ATE RMSE[cm]$\downarrow$ for different sequences. }
    \label{tab:replica_tracking}
\end{table}

\begin{table}[t]
    \centering
    \scriptsize
    \renewcommand{\arraystretch}{1.0}
    \resizebox{0.9\columnwidth}{!}
    {
    \begin{tabular}{l|ccccc|c}
    \hline 
    Method                                                &    0000   &   0059    &   0169    &   0181    &   0207    &\textbf{Avg.}\\
    \hline
    \multirow{1}{*}{NICE-SLAM~\cite{zhu2022nice}}         &      12.0 &      14.0 &      10.9 &      13.4 &      6.2  &      11.30 \\[0.8pt]
    \multirow{1}{*}{Vox-Fusion~\cite{yang2022vox}}        &      16.6 &      24.2 &      27.3 &      23.3 &      9.4  &      20.14 \\[0.8pt]
    \multirow{1}{*}{Point-SLAM~\cite{sandstrom2023point}} &      10.2 &      7.8  &      22.2 &      14.8 &      9.5  &      12.90 \\[0.8pt]
    \multirow{1}{*}{DNS-SLAM~\cite{li2024dns}}            &      12.1 &      5.5  &      35.6 &      10.0 &      6.4  &      13.92 \\[0.8pt]
    \multirow{1}{*}{SNI-SLAM~\cite{zhu2024sni}}           &      6.9  &      7.4  &      -    &      -    &      4.7  &      -     \\[0.8pt]
    \multirow{1}{*}{SplaTAM~\cite{keetha2024splatam}}     &      12.8 &      10.1 &      12.1 &      11.1 &      7.5  &      10.72 \\[0.8pt]
    \multirow{1}{*}{MonoGS~\cite{matsuki2024gaussian}}    &      9.8  &      6.4  &      10.7 &      23.8 &      8.1  &      11.76 \\[0.8pt]
    \multirow{1}{*}{SGS-SLAM~\cite{li2024sgs}}            &      12.6 &      14.0 &      17.6 &      13.0 &      8.0  &      13.04 \\[0.8pt]
    \multirow{1}{*}{LoopSplat*~\cite{zhu2025_loopsplat}}  &       6.2 &      7.1  &      10.6 &      8.5  &      6.6  &      6.48 \\[0.8pt]
    \multirow{1}{*}{GSFF-SLAM(Ours)}                      &      9.6 &       7.8  &      9.6  &      21.9 &      5.6  &      10.90 \\[2pt]

    \hline
    \end{tabular}
    }
    \caption{Comparison of tracking performance on ScanNet dataset~\cite{dai2017scannet}(ATE RMSE$\downarrow$[cm]). LoopSplat* is a 3DGS-based method using loop closure.}
    \label{tab:scannet_tracking}
\end{table}

\begin{table}[t]
    \centering
    \scriptsize
    \renewcommand{\arraystretch}{1.2}
    \resizebox{0.9\columnwidth}{!}{
    \begin{tabular}{l|ccccc|c}
    \hline 
    \multirow{2}{*}{Method}                               &fr1/       &fr1/       &fr1/       &fr2/       &fr3/       &\multirow{2}{*}{\textbf{Avg.}}\\
                                                          &desk1      &desk2      &room       &xyz        &office        & \\
    \hline
    \multirow{1}{*}{NICE-SLAM~\cite{zhu2022nice}}         &      4.26 &      4.99 &      34.49&      6.19 &      3.87 &      10.76 \\[0.8pt]
    \multirow{1}{*}{Vox-Fusion~\cite{yang2022vox}}        &      3.52 &      6.00 &      19.53&      1.49 &      26.01&      11.31 \\[0.8pt]
    \multirow{1}{*}{Point-SLAM~\cite{sandstrom2023point}} &      4.34 &      4.54 &      30.92&      1.31 &      3.48 &      8.92  \\[0.8pt]
    \multirow{1}{*}{SNI-SLAM~\cite{zhu2024sni}}           &      2.56 &      4.35 &      11.46&      1.12 &      2.27 &      4.35  \\[0.8pt]
    \multirow{1}{*}{SplaTAM~\cite{keetha2024splatam}}     &      3.35 &      6.54 &      11.13&      1.24 &      5.16 &      5.48  \\[0.8pt]
    \multirow{1}{*}{MonoGS~\cite{matsuki2024gaussian}}    &      1.59 &      7.03 &      8.55 &      1.44 &      1.49 &      4.02  \\[0.8pt]
    \multirow{1}{*}{LoopSplat*~\cite{zhu2025_loopsplat}}  &      2.08 &      3.54 &      6.24 &      1.58 &      3.22 &      3.33  \\[0.8pt]
    \multirow{1}{*}{ORB-SLAM2$\star$~\cite{mur2017orb}}   &      1.6  &      2.2  &      4.7  &      0.4  &      1.0  &      1.98  \\[0.8pt]
    \multirow{1}{*}{GSFF-SLAM(Ours)}                      &      1.61 &      7.30 &      7.29 &      1.61 &      1.96 &      3.95  \\[0.8pt]
    \hline
    \end{tabular}
    }
    \caption{Comparison of tracking performance on TUM-RGBD dataset~\cite{sturm2012benchmark}(ATE RMSE$\downarrow$[cm]). ORB-SLAM2$\star$ is a feature-based SLAM method using loop closure.}
    \label{tab:tum_tracking}
\end{table}

\begin{figure*}[!t]
   \centering
   \includegraphics[width=\linewidth]{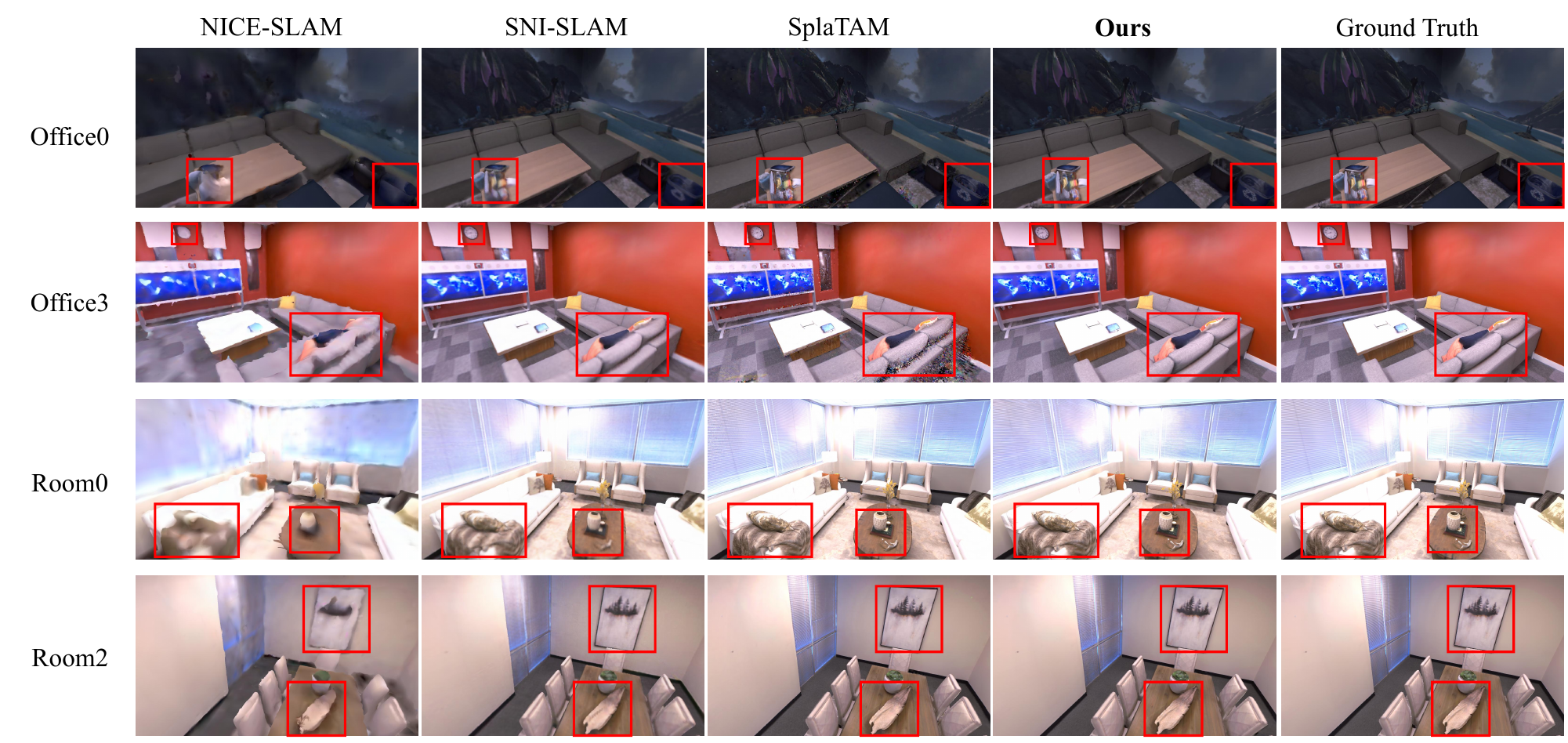}
   \caption{Qualitative comparison on rendering quality of baseline and our method. We select 4 scenes of Replica dataset and highlighted the differences with red color boxes. }
   \label{fig:replica_render}
\end{figure*}

\begin{table}[t]
    \centering
    \renewcommand{\arraystretch}{1.5}
    \resizebox{0.9\columnwidth}{!}{
    \begin{tabular}{l|ccc|ccc|ccc}
    \hline
    \multirow{1}{*}{Dataset}    & \multicolumn{3}{c|}{\textbf{Replica}}  & \multicolumn{3}{c|}{\textbf{Scannet}}   & \multicolumn{3}{c}{\textbf{TUM-RGBD}}   \\
    \hline
    \multirow{1}{*}{Method}     & PSNR$\uparrow$ & SSIM$\uparrow$ & LPIPS$\downarrow$ &PSNR$\uparrow$ & SSIM$\uparrow$ & LPIPS$\downarrow$ & PSNR$\uparrow$ & SSIM$\uparrow$ & LPIPS$\downarrow$ \\
    \hline
    \multirow{1}{*}{NICE-SLAM\cite{zhu2022nice}}            &     24.42  &     0.892  &      0.233 &     17.54  &     0.621 &     0.548 &     14.86  &     0.614  &     0.441     \\[0.8pt]  
    \multirow{1}{*}{Vox-Fusion\cite{yang2022vox}}           &     24.41  &     0.801  &      0.236 & \rd 18.17  &     0.673 &     0.504 &     16.46  &     0.677  &     0.471     \\[0.8pt]  
    \multirow{1}{*}{DNS-SLAM\cite{li2024dns}}               &     20.91  &     0.758  &      0.208 &     10.17  &     0.551 &     0.725 &     -      &     -      &     -         \\[0.8pt]  
    \multirow{1}{*}{Point-SLAM\cite{sandstrom2023point}}    &     35.05  & \fs 0.975  &      0.124 &     10.34  &     0.557 &     0.748 &     16.62  &     0.696  &     0.526     \\[0.8pt]  
    \multirow{1}{*}{SplaTAM\cite{keetha2024splatam}}        & \rd 33.55  & \nd 0.971  & \rd  0.090 & \fs 19.04  & \nd 0.699 & \fs 0.367 & \fs 22.25  & \fs 0.878  & \fs 0.181     \\[0.8pt]  
    \multirow{1}{*}{MonoGS\cite{matsuki2024gaussian}}       & \nd 35.77  &     0.951  & \nd  0.067 &     17.11  & \fs 0.734 &     0.581 & \rd 18.60  & \rd 0.713  & \rd 0.334     \\[0.8pt]  
    \multirow{1}{*}{SGS-SLAM\cite{li2024sgs}}               &     33.53  &     0.966  & \rd  0.090 &     18.07  &     0.695 & \rd 0.417 & -          &     -      &     -         \\[0.8pt]
    \multirow{1}{*}{Ours}                        & \fs 38.67  & \rd 0.974  & \fs  0.035 & \nd 18.90  & \rd 0.697 & \nd 0.391 & \nd 20.57  & \nd 0.736  & \nd 0.311     \\[0.8pt] 
    \hline
    \end{tabular}  
    }
    \caption{Quantitative comparison of rendering performance on the Replica~\cite{straub2019replica}, ScanNet~\cite{dai2017scannet}, and TUM-RGBD~\cite{sturm2012benchmark} datasets. The best results are highlighted as \colorbox{tabfirst}{\textbf{first}}, \colorbox{tabsecond}{second}, and \colorbox{tabthird}{third}.}
    \label{tab:renderingperf}
\end{table}

\textbf{Tracking Accuracy}.
On the Replica dataset, we evaluate the tracking performance of our method compared to the state-of-the-art methods. As shown in Table~\ref{tab:replica_tracking}, our method achieves the best tracking accuracy with an average ATE RMSE of 0.311cm. In the high-quality synthetic environments with accurate depth information and minimal motion blur, our method achieves submillimeter tracking accuracy that is comparable to other state-of-the-art 3DGS-based methods, while consistently outperforming NeRF-based methods. Compared to the NeRF-based SLAM, our method reduces the tracking error by more than 25\% from 0.456cm to 0.311cm. Compared to the recent SplaTAM and MonoGS, which achieve average ATE RMSE of 0.384cm and 0.321cm respectively, our method reduces the tracking error by 19.0\% and 3.1\%. 
On real-world datasets, Table~\ref{tab:scannet_tracking} and Table~\ref{tab:tum_tracking} present the quantitative comparison of tracking accuracy. Our method achieves performance with average ATE RMSE of 10.90cm on ScanNet and 3.95cm on TUM-RGBD. In the ScanNet dataset, we observe certain limitations of our approach. Specifically, in scene0000 and scene0181, the tracking performance exhibits significant degradation. For scene0000, which represents the longest sequence in our evaluation, we observe gradual drift in trajectory estimation after approximately 3000 frames, primarily due to the absence of loop closure detection. Meanwhile, scene0181 presents additional challenges due to significant motion blur, which adversely affects the mapping quality. These observations highlight potential areas for future improvement, particularly in handling extended sequences and scenes with degraded image quality. On the TUM-RGBD dataset, we conduct comprehensive experiments to further verify the robustness of our approach. Our method achieves competitive performance with an average ATE RMSE of 3.95cm, ranking among all learning-based methods. It is noteworthy that our algorithm performs poorly in the scene with motion blur, fr1/desk2, where the ATE RMSE of our method is 7.30 cm.
In comparison, the NeRF-based SNI-SLAM~\cite{zhu2024sni} achieve ATE RMSE values of 4.35 cm, respectively.

\noindent\textbf{Rendering Quality}.
We report the rendering performance in Table\ref{tab:renderingperf} on the three datasets, our method achieves the highest PSNR and LPIPS scores in Replica dataset, indicating that our method can generate high-quality rendered images. As shown in Figure \ref{fig:replica_render}, we show the qualitative comparison of rendering quality between the baseline and our method on Replica dataset and our method achieves better texture details in the scene reconstruction. On real-world datasets, our method behaves robust in rendering quality, which is consistent with the quantitative results.
While performance on real-world datasets lags behind synthetic data, this underscores the challenge of reconstructing high-quality scenes from noisy and motion-blurred sequences.

\begin{figure*}[t]
   \centering
   \includegraphics[width=\linewidth]{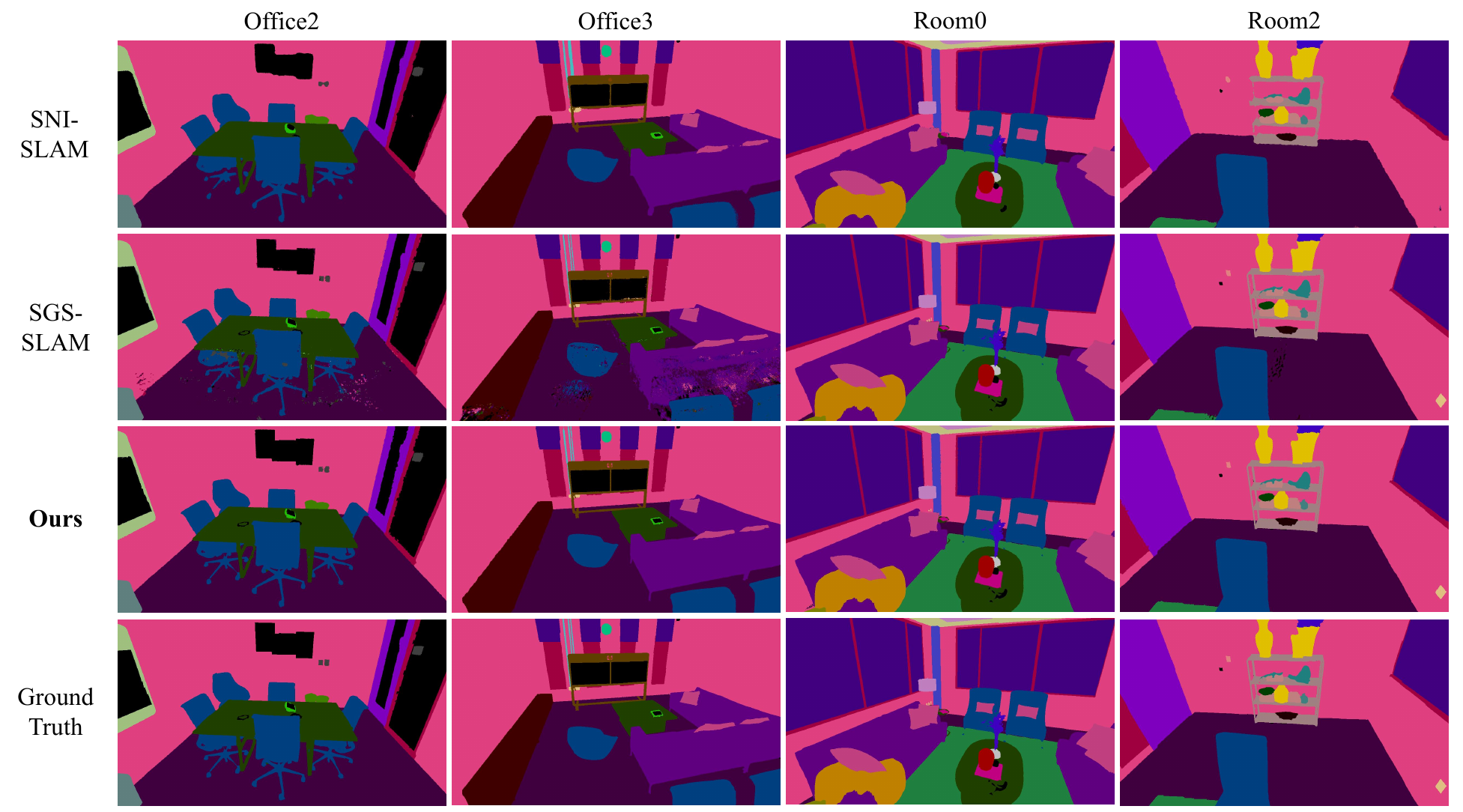}
   \caption{Qualitative comparison of semantic reconstruction performance using ground truth labels on the Replica dataset~\cite{straub2019replica}.}
   \label{fig:ground truth segmentation}
\end{figure*}

\begin{figure*}[t]
   \centering
   \includegraphics[width=\linewidth]{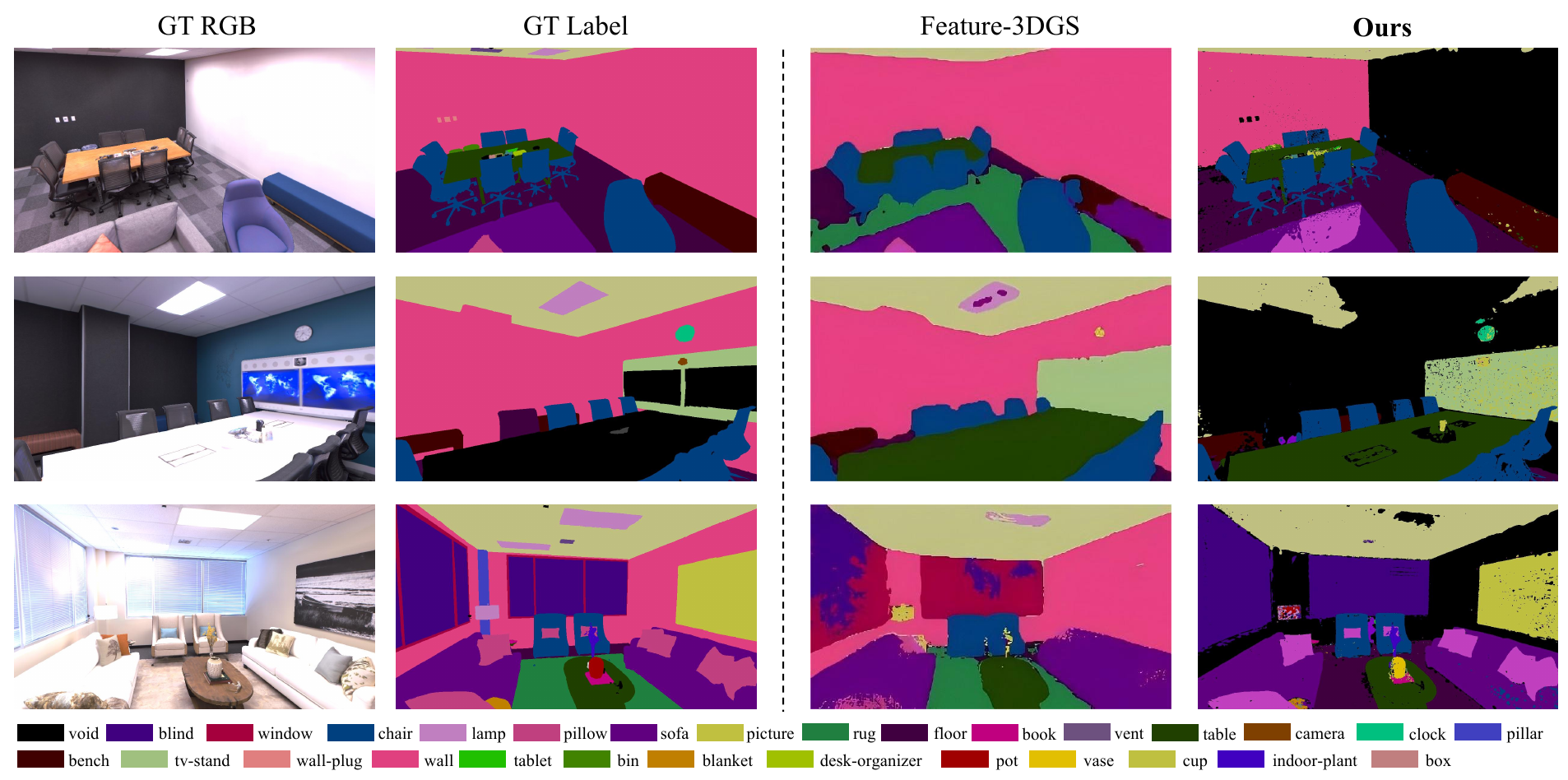}
   \caption{Qualitative comparison of semantic reconstruction performance using noisy textual labels on the Replica dataset~\cite{straub2019replica}. We merge semantically similar objects with high 3D spatial overlap, such as windows and blinds, rugs and floors.}
   \label{fig:noisy label segmentation}
\end{figure*}

\subsection{Evaluation of Semantic Reconstruction}

\begin{table}[]
    \centering
    \renewcommand{\arraystretch}{1.2}
    \resizebox{0.9\columnwidth}{!}{
    \begin{tabular}{ccccccccccc}
    \toprule
    \textbf{Method}         & \textbf{Metric}                       &\textbf{O0} &\textbf{O1} &\textbf{O2} &\textbf{O3} &\textbf{O4} &\textbf{R0} &\textbf{R1} &\textbf{R2} &\textbf{Avg.}  \\[0.8pt] 
    \midrule
                                                            & Acc$\uparrow$   & \rd 98.89  &    -       &    -       &     -      &     -      &     97.76  &     98.50  & \rd 98.76  &     98.47 \\[0.8pt]  
    \multirow{-2}{*}{\makecell{NIDS-\\SLAM}~\cite{haghighi2023neural}}                & mIoU$\uparrow$  &     85.94  &    -       &    -       &     -      &     -      &     82.45  &     84.08  &     76.99  &     82.37 \\[0.8pt]  \hdashline \noalign{\vskip 2pt}
                                                            & Acc$\uparrow$   &     98.27  &     97.79  &     98.36  &     97.20  &     90.59  &     97.85  &     97.62  &     98.27  &     96.81 \\[0.8pt]  
    \multirow{-2}{*}{\makecell{DNS-\\SLAM}~\cite{li2024dns}}                 & mIoU$\uparrow$  &     84.10  &     82.75  &     73.33  &     74.13  &     64.74  &     85.81  &     85.80  &     77.86  &     78.56 \\[0.8pt]  \hdashline \noalign{\vskip 2pt}
                                                            & Acc$\uparrow$   &     98.73  & \rd 98.98  & \rd 98.93  & \nd 98.58  & \nd 99.00  & \rd 98.32  & \rd 98.63  &     98.73  & \rd 98.73 \\[0.8pt] 
    \multirow{-2}{*}{\makecell{SNI-\\SLAM}~\cite{zhu2024sni}}                 & mIoU$\uparrow$  & \rd 87.66  & \rd 85.13  & \rd 84.00  & \rd 79.39  & \rd 78.11  & \rd 87.22  & \rd 88.11  & \rd 84.62  & \rd 84.62 \\[0.8pt] \hdashline \noalign{\vskip 2pt}
                                                            & Acc$\uparrow$   & \nd 99.52  & \fs 99.59  & \nd 99.18  & \rd 98.29  & \rd 98.77  & \nd 99.12  & \nd 99.36  & \nd 99.36  & \nd 99.14 \\[0.8pt] 
    \multirow{-2}{*}{\makecell{SGS-\\SLAM}~\cite{li2024sgs}}                 & mIoU$\uparrow$  & \nd 92.29  & \fs 95.25  & \nd 91.62  & \nd 87.25  & \nd 90.34  & \nd 92.47  & \nd 92.43  & \nd 92.83  & \nd 91.81 \\[0.8pt] \hdashline \noalign{\vskip 2pt}
                                                            & Acc$\uparrow$   & \fs 99.53  & \nd 99.46  & \fs 99.55  & \fs 99.50  & \fs 99.11  & \fs 99.16  & \fs 99.55  & \fs 99.56  & \fs 99.41 \\[0.8pt] 
    \multirow{-2}{*}{Ours}                                  & mIoU$\uparrow$  & \fs 95.12  & \nd 93.44  & \fs 94.96  & \fs 94.74  & \fs 93.97  & \fs 95.20  & \fs 96.90  & \fs 95.98  & \fs 95.03 \\[0.8pt] 
    \bottomrule
    \end{tabular}
    }
    \caption{Quantitative comparison of GSFF-SLAM with existing semantic SLAM methods for semantic segmentation metrics \textit{Acc(\%)} and \textit{mIou(\%)}  on the Replica dataset~\cite{straub2019replica}.}
    \label{tab:replica_seg_metrics}
\end{table}

\begin{table}[]
    \centering
    \renewcommand{\arraystretch}{1.1}
    \resizebox{0.7\columnwidth}{!}{
    \begin{tabular}{lccc}
    \hline
    Method & mIoU(\%)$\uparrow$ & Inference(fps)$\uparrow$ & Time(min)$\downarrow$  \\
    \hline
    SNI-SLAM~\cite{zhu2024sni}      & 84.62 & 0.87  & 132 \\
    SGS-SLAM~\cite{li2024sgs}       & 91.81 & 392   & 234 \\
    Ours (w/ speedup)                & 90.54 & 19.2  & 45  \\
    Ours                            & 95.03 & 15.8  & 114 \\
    \hline
    \end{tabular}}
    \caption{Performance of semantic SLAM on Replica dataset~\cite{straub2019replica} compared to SNI-SLAM~\cite{zhu2024sni} and SGS-SLAM~\cite{li2024sgs}. }
    \label{tab:compare_pipe}
\end{table}

\noindent\textbf{Ground Truth Supervision.}  
As shown in Table\ref{tab:replica_seg_metrics}, our method outperforms the all baseline methods in segmentation metrics of Replica dataset by using the ground truth semantic labels. Specifically, our approach demonstrates superior semantic segmentation accuracy with an average pixel accuracy of 99.41\% and mIoU of 95.03\% across all sequences, significantly outperforming previous methods. Compared to the recent SNI-SLAM, our method improves the accuracy by 0.68\% and notably boosts the mIoU by 10.41\%. 
We also show the qualitative comparison of rendering quality between the baseline and our method in Figure \ref{fig:ground truth segmentation}. It can be observed that SNI-SLAM suffers from blurry boundaries, poor recognition of small objects, and loss of high-frequency details. SGS-SLAM exhibits some noise in the reconstructed local regions. In contrast, our method produces clearer semantic boundaries while preserving small object information, showing overall superior stability. 
These results further demonstrate that, due to the noise and uncertainty inherent in the reconstruction process, incorporating semantic supervision signals and semantic loss to improve tracking accuracy is not particularly crucial. The adoption of this separate gradient design makes our algorithm framework more flexible, and the semantic supervision learning process more stable.
As shown in Table \ref{tab:compare_pipe}, we compare the performance of SNI-SLAM in semantic segmentation, inference rendering speed, and runtime. Our optimized semantic rendering pipeline can achieve a semantic rendering speed of up to 19.2 fps. When using \(\tau_{\text{thresh}}=0.8\) and \(\rho_{\text{pc}}=1/64\), our method with speedup still outperforms SNI-SLAM in semantic segmentation performance, but achieves a 2.9\(\times\) improvement in runtime. In contrast, SGS-SLAM does not directly render features but instead reconstructs them through a label-to-RGB conversion strategy, achieving the highest rendering speed.

\noindent\textbf{Noisy Textual Label Supervision.}  
Figure \ref{fig:noisy label segmentation} shows the visualization results using noisy textual priors. Compared to the Feature-3DGS~\cite{zhou2024feature} method, which utilizes LSeg~\cite{li2022language} as the base model, our online rendering pipeline demonstrates that sparse priors with lower noise and accurate edges significantly enhance semantic reconstruction in detailed regions, as opposed to using noisy dense semantic priors. Furthermore, it is noteworthy that some unlabelled objects, represented in black, further validating the feasibility of incorporating the foundation model into our semantic reconstruction pipeline for downstream tasks. However, sparse detection may result in certain objects being contaminated by surrounding areas due to low detection rates, with the ceiling light and vent shown in the Figure \ref{fig:noisy label segmentation} being a typical example.

\subsection{Ablation Study}

\begin{table}[]
    \centering
    \renewcommand{\arraystretch}{1.2}
    \resizebox{0.9\columnwidth}{!}{
    \begin{tabular}{cc|ccccc}
    \hline
    $\rho_{\text{pc}}$  & $\tau_{\text{thresh}}$  & RMSE(cm)$\downarrow$ & PSNR(dB)$\uparrow$ & mIoU(\%)$\uparrow$ & Mem(MB)$\downarrow$ & Time(min)$\downarrow$ \\
    \hline
     1/64 & 0.8  & 0.582   & 36.41  & 90.54    & 5281    & 45        \\
     1/64 & 0.95 & 0.495   & 37.78  & 93.68    & 9088    & 63        \\
     1/16 & 0.8  & 0.396   & 37.64  & 92.74    & 8360    & 64        \\
     1/16 & 0.95 & 0.390   & 38.67  & 95.03    & 12062   & 114       \\
    \hline
    \end{tabular}}
    \caption{Ablation study on Replica dataset using ground truth. We studied the effects of co-visibility overlap threshold $\tau_{\text{thresh}}$ and point cloud downsampling rate $\rho_{\text{pc}}$ on the system.}
    \label{tab:ablation}
\end{table}

We conducted an ablation study to examine the effects of two key hyperparameters: point cloud downsampling rate \(\rho_{\text{pc}}\) and co-visibility overlap threshold \(\tau_{\text{thresh}}\) on system performance, as shown in Table \ref{tab:ablation}.
Our analysis reveals that \(\rho_{\text{pc}}\) plays a crucial role in determining tracking accuracy.
This suggests that denser point clouds provide more detailed information, which is beneficial for tracking accuracy. 
On the other hand, \(\tau_{\text{thresh}}\) significantly affects semantic reconstruction performance. A higher threshold improves the mIoU by allowing for better alignment of semantic features, which leads to more accurate semantic reconstruction. This is evident in the marked improvement in mIoU when increasing \(\tau_{\text{thresh}}\) from 0.8 to 0.95. 
both hyperparameters have minimal impact on rendering quality, but they do affect runtime.
From the results, we observe that the hyperparameter settings in the first row (\(\rho_{\text{pc}} = 1/64, \tau_{\text{thresh}} = 0.8\)) strike a balance between memory usage, runtime, and accuracy. These settings minimize memory requirements while maintaining competitive tracking performance and rendering quality, making them the optimal choice for resource-constrained deployment scenarios.

\section{Conclusion}



We propose GSFF-SLAM, a dense semantic SLAM system that employs 3D Gaussian Splatting as the map representation and incorporates semantic features for efficient semantic map rendering. We leverage the overlap between consecutive frames to select keyframes and independently optimize the gradients of features. Extensive experiments demonstrate the robustness of our method in tracking and mapping, as well as the accuracy of semantic edges. We verify the feasibility of semantic reconstruction using noisy sparse signals, opening up a pathway for online semantic reconstruction in previously unseen scenes. Future work will focus on extending GSFF-SLAM to dynamic environments and further improving pipeline efficiency.



\bibliographystyle{unsrt}  
\bibliography{main}  

\end{document}